\def\year{2021}\relax
\documentclass[letterpaper]{article} 
\usepackage{aaai21}  
\usepackage{times}  
\usepackage{helvet} 
\usepackage{courier}  
\usepackage[hyphens]{url}  
\usepackage{graphicx} 
\usepackage{natbib}  
\usepackage{caption} 
\usepackage{multirow}
\title{Efficient Object-Level Visual Context Modeling for Multimodal Machine Translation: Masking Irrelevant Objects Helps Grounding}
\author{
    Dexin Wang,
    Deyi Xiong\\
}
\affiliations{
    College of Intelligence and Computing, Tianjin University, Tianjin, China, 300350\\

    \{dyxiong, dexinwang\}@tju.edu.cn

}

\begin{document}

\maketitle

\begin{abstract}
Visual context provides grounding information for multimodal machine translation (MMT). However, previous MMT models and probing studies on visual features suggest that visual information is less explored in MMT as it is often redundant to textual information.
In this paper, we propose an object-level visual context modeling framework (OVC) to efficiently capture and explore visual information for multimodal machine translation.
With detected objects, the proposed OVC encourages MMT to ground translation on desirable visual objects by masking irrelevant objects in the visual modality. We equip the proposed with an additional object-masking loss to achieve this goal. The object-masking loss is estimated according to the similarity between masked objects and the source texts so as to encourage masking source-irrelevant objects. Additionally, in order to generate vision-consistent target words, we further propose a vision-weighted translation loss for OVC. Experiments on MMT datasets demonstrate that the proposed OVC model outperforms state-of-the-art MMT models and analyses show that masking irrelevant objects helps grounding in MMT.

\end{abstract}

\section{Introduction}

Multimodal Machine Translation aims at translating a sentence paired with an additional modality (e.g. audio modality in spoken language translation or visual modality in image/video-guided translation) into the target language \cite{multi30k}, where the additional modality, though closely semantically related to the text, provides an alternative and complementary view to it.
By contrast to text-only neural machine translation (NMT), MMT characterizes with the assumption that the additional modality helps improve translation by either grounding the meaning of the text or providing multimodal context information \cite{emergent}.
Hence, MMT exhibits pronounced reliance on language-vision/speech interaction.\footnote{In this paper, we focus on multimodal machine translation with both visual and textual modalities.}

However, effectively integrating visual information and language-vision interaction into machine translation has been regarded as a big challenge  \cite{agreement} for years since Multi30K \cite{multi30k} is proposed as a benchmark dataset for MMT.
 \begin{figure}[t]
 \centering
 \includegraphics[width=1.0\columnwidth]{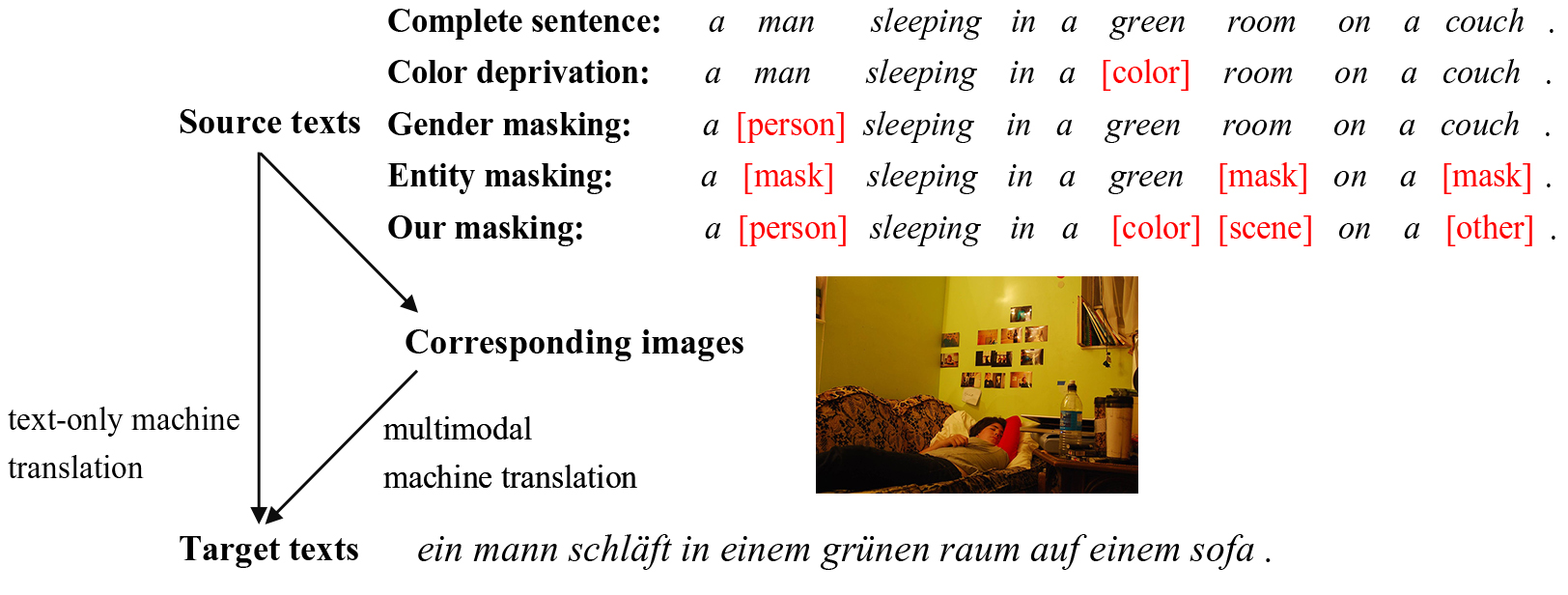} 
 \caption{Word masking in multimodal machine translation.}
 \label{fig1}
 \end{figure}
Many previous MMT studies on Multi30K, which exploit complete source texts during both training and inference, have found that visual context is needed only in special cases, e.g., translating sentences with incorrect or ambiguous source words, by both human and machine translation, and is hence marginally beneficial to multimodal machine translation \cite{sheffield, distilling}.

In this paper, we hypothesize that visual context can be efficiently exploited to enhance MMT, instead of being ignored as a redundant input, from three aspects as follows:
\begin{itemize}
\item \emph{Source texts processing and encoding}: In most cases, source texts provide sufficient information for translation, which makes visual context redundant. Therefore, weakening the input signal from the textual modality may force MMT to pay more attention to the visual modality.
\item \emph{Visual feature learning tailored for translation}: Not all parts in visual images are useful for translation. Learning visual features that are not only linked but also complementary to source texts is desirable for MMT.
\item \textit{Target words generation and decoding}: Visual representations can be used to not only initialize the decoder \cite{vag-nmt} but also guide target word prediction (e.g., rewarding target prediction consistent with visual context).
\end{itemize}

Regarding the first aspect, we have witnessed that pioneering efforts \cite{probing,distilling}, different from previous methods, mask specific words (e.g. gender-neutral words) in source texts, forcing MMT to distill visual information into text generation, as shown in Figure \ref{fig1}.
In addition to the source text masking, in this paper, we attempt to explore all the three aforementioned aspects in a unified framework for MMT.
Specifically, we propose an efficient object-level visual context modeling framework (OVC) to capture desirable visual features and to reward vision-consistent target predictions for MMT.
In this framework,  we first detect a bag of objects from images. Inspired by the word masking method in source texts \cite{probing}, we also encourage OVC to mask visual objects that are not relevant to source texts by computing object-text similarity in a preprocessing step.
For this, we propose an object-masking loss that calculates the cross-entropy loss difference between original translation and translations generated with the relevant-object-masked image vs. irrelevant-object-masked image. This is to reward masking irrelevant objects in visual context while masking relevant objects is penalized.

In order to force the decoder to generate vision-consistent target words, we change the traditional cross-entropy translation loss into a vision-weighted loss in OVC, which tends to reward the generation of vision-related words or rare but vision-consistent words.

 To examine the effectiveness of the proposed OVC in visual feature learning, we test OVC against the baselines in both standard and source-degradation setting with word masking as shown in Figure \ref{fig1}.

The contributions of this work can be summarized as follows:
\begin{itemize}
\item We propose a new approach to MMT, which masks both objects in images and specific words in source texts for better visual feature learning and exploration.
\item We propose two additional training objectives to enhance MMT: an object-masking loss to penalize undesirable object masking and a vision-weighted translation loss to guide the decoder to generate vision-consistent words.
\item We conduct experiments and in-depth analyses on existing MMT datasets, which demonstrate that our model can outperform or achieve competitive performance against the-state-of-the-art MMT models.
\end{itemize}

\section{Related Work}

\subsection{MMT without Text Masking}
Since the release of the Multi30K dataset, a variety of different approaches have been proposed for multimodal machine translation.
Efforts for the MMT modeling mechanism can be categorized into RNN-based sequence-to-sequence models and attention-based ones.
\citeauthor{imagination} \shortcite{imagination} and \citeauthor{LIUMCVC} \shortcite{LIUMCVC} employ GRU/LSTM-based encoder-decoder models to encode source texts and integrate a single image vector into the model. The image vector is either used to initialize the encoder or decoder \cite{vag-nmt, distilling} or to fuse with word embeddings in the embedding layer of the encoder \cite{LIUMCVC}.
Attention-based sequence-to-sequence approaches have been proposed for MMT \cite{huang}, which compute either spatially-unaware image-to-texts attention \cite{zhang} or spatially-aware object-to-text to capture vision-text interaction so as to enhance the encoder and decoder of MMT \cite{agreement}.

We also have witnessed two proposed categories for MMT from the perspective of cross-modal learning approaches, which either explicitly transform visual features and textual embeddings from one modality to the other at both training and inference \cite{LIUMCVC, gmmt}, or implicitly align the visual and textual modalities to generate vision-aware textual features at training.
 Unlike the explicit approaches, the implicit cross-modal learning methods do not require images as input at inference, taking the image features as latent variables across different languages \cite{imagination, latent, embedding}, which also serves as a latent scheme for unsupervised MMT \cite{emergent}.
Despite of the success of plenty of models on Multi30K, an interesting finding is that the visual modality is not fully exploited and only marginally beneficial to machine translation \cite{LIUMCVC, distilling}.

\subsection{Text-Masked MMT}

To probe the real need for visual context in MMT, several researchers further explore new settings where visual features are not explicitly expressed by source texts on purpose.
 In other words, specific source words that are linked to visual features are purposely masked.
In particular, \citeauthor{distilling} \shortcite{distilling} focus on three major linguistic phenomena and mask ambiguous, inaccurate and gender-neutral (e.g., player) words in source texts on Multi30K.
 Their experiment results suggest that the additional visual context is important for addressing these uncertainties.
\citeauthor{probing} \shortcite{probing} propose more thoroughly masked schemes on Multi30K by applying color deprivation, whole entity masking and progressive masking on source texts.
They find that MMT is able to integrate the visual modality when the available visual features are complementary rather than redundant to source texts.

Although masking source words forces MMT models to pay more attention to and therefore exploit the visual modality for translation, there is a big performance gap between the standard setting (without text masking) and source-degradation setting (purposely masking specific words). For example, in the experiments reported by \citeauthor{distilling} \shortcite{distilling}, the best METEOR on WMT 2018 MMT EN-DE test set for the standard setting is 46.5 while the highest METEOR score for the source-degradation setting is only 41.6.
Although specific words are masked in source texts, visual features that are semantically linked to these words are available in the visual modality provided for MMT. This indicates that the visual modality is not fully exploited by current MMT models even though the available information is complementary to source texts.

\begin{figure*}[t]
    \centering
        \includegraphics[width=2.\columnwidth]{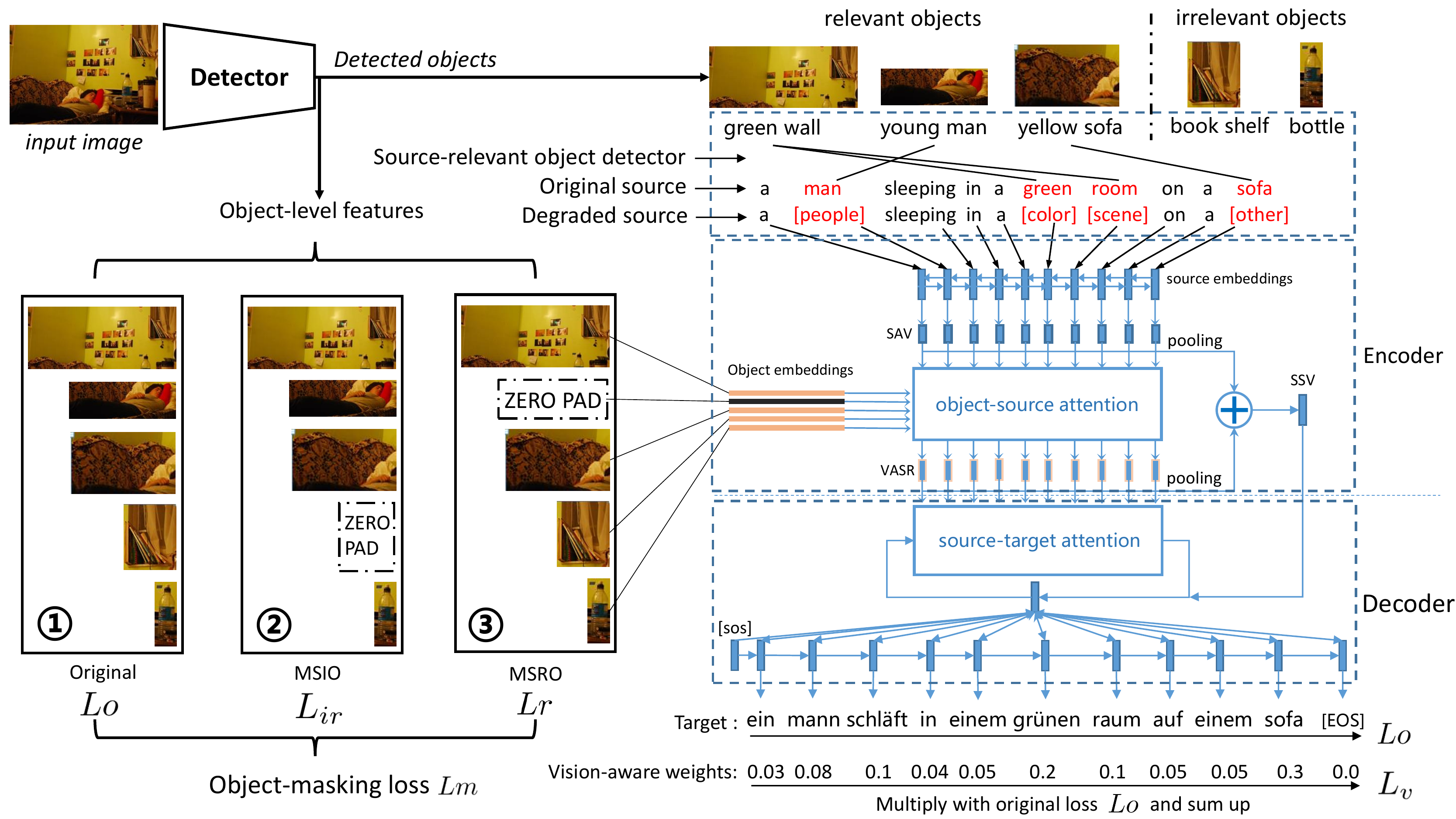} 
        \caption{The architecture of the proposed OVC framework. MSIO: masking source-irrelevant objects. MSRO: masking source-relevant objects. SAV denotes source annotation vectors. VASR is the vision-aware source representation of the source sentence.}
    \label{fig2}
\end{figure*}

\section{Efficient Object-Level Visual Context Modeling}

In this section, we elaborate the proposed OVC model.
The backbone of the model is a GRU-based encoder-decoder neural network with two multihead attention layers that model the attention between source tokens and detected objects in the input image as well as the attention between target and source tokens.
The architecture of OVC is shown in Figure \ref{fig2}.
The source input to OVC can be either an original source sentence or the degradation of the source sentence (see Section `Experiment' for more details on how we degrade source sentences by masking specific words).
The visual modality is integrated into the model through the object-source multihead attention, which is also explored in two additional training objectives: the object-masking loss and vision-weighted translation loss.

\subsection{Encoder}
The encoder of OVC consists of a bidirectional GRU module and an object-source attention layer that performs the fusion of textual and visual modality.
The inputs to the encoder include token embeddings of source texts and object-level visual features from the paired image.
Let $W_s^n=\{w_s^1, w_s^2,...,w_s^n\}$ denotes the token embedding matrix of the source sentence, where $n$ is the number of tokens.
The object-level features are a set of vector embeddings of objects detected by a pre-trained object detector.
Each detected object is labeled with its predicted object category and attribute (e.g., ``young man", ``green wall").
In our case, we use Resnet101 \cite{resnet} as the object detector which compresses each object into a 2048-dimension vector.
We denote the object embedding matrix as $O^m=\{o^1, o^2, ..., o^m\}$, where $m$ is the number of all detected objects.
During training, some objects from the paired image are randomly selected and masked, which we'll discuss in the following subsection in details.
The representation for a masked object is set to a zero vector.

The bidirectional GRU transforms the sequence of source token embeddings into a sequence of annotation vectors (SAV):
\begin{equation}
    H_s^n = (h_s^1, h_s^2, ..., h_s^n)
\end{equation}
We then adopt a multihead attention layer over $H_s^n$ and $O^m$ to obtain a vision-aware source representation (VASR) as follows:
\begin{equation}
\textmd{VASR} = \textmd{MultiHead}_1(H_s^n, O^m, O^m)
\end{equation}
where MultiHead(Q, K, V) is a multihead attention function taking a query matrix Q, a key matrix K, and a value matrix V as inputs.
After that, we aggregate VASR and $H_s^n$ into a mixed-modality source sentence vector (SSV) by applying average-pooling (AP) on both VASR and $H_s^n$ to get two separate vectors and then adding the two vectors as follows:
\begin{equation}
\textmd{SSV} = \textmd{AP}(\textmd{VASR})  + \textmd{AP}(H_s^n)
\end{equation}

\subsection{Decoder}
The decoder of OVC also consists of a multihead attention layer to compute source-target attention and a GRU module to update hidden states.
SSV is fed into the GRU layer to initialize the decoder as follows:
\begin{equation}
    H^0_t = \textmd{GRU}(w_t^{\textmd{[sos]}}, \textmd{SSV})
\end{equation}
where $w_t^{[sos]}$ is the embedding of the start symbol.
At each time step, the multihead attention layer computes the source-target attention as follows:
\begin{equation}
    T^{i+1} = \textmd{MultiHead}_2(H_t^{i},\textmd{VASR}, \textmd{VASR})
\end{equation}
where $H_t^i$ is the hidden state at time step $i$ of the decoder.
The GRU module aggregates previous word embedding and $T^{i+1}$ to update the hidden state as follows:
\begin{equation}
    H^{i+1} = \textmd{GRU}(w_t^{i}, T^{i+1})
\end{equation}
where $w_t^i$ denotes the embedding of the $i$-th target word.
Finally, we project $H_t$ into logit vectors for target word prediction over the vocabulary.

\subsection{Training Objectives}
In order to facilitate our model to capture the deep interaction between the textual and visual modality, in OVC, we propose two additional translation objectives to better integrate visual features into MMT: an object-masking loss and a vision-weighted translation loss.
\label{object-masking}
\subsubsection{Object-Masking Loss.}
The object-masking loss (denoted as $L_m$) is to optimize MMT to discriminate good grounding of source tokens to the visual modality from bad grounding by telling the model the difference between masking source-relevant objects and masking those irrelevant.
If an object is masked, the corresponding $o^i$ is set to a zero vector.
Specifically, the goals of using this objective are two-folds:
\begin{itemize}
    \item forcing the model to penalize masking objects on which source words (or tags in degraded source sentences) can be grounded.
    \item rewarding masking schemes where irrelevant objects are masked so as to avoid the negative impact from them.
\end{itemize}

Before we define the object-masking loss, let's discuss how we can detect source-relevant objects from those irrelevant.
Generally, we compute the degree of the relevance of an object to the source sentence by semantic similarity with the aid of a pre-trained language model.\footnote{In this paper, we use multilingual-cased-base-BERT which is only used to detect source-relevant objects during preprocessing. Other word embedding methods can be also used here for computing similarity.}
In particular, we first compute a cosine similarity matrix (denoted as $S^{m*n}$) for all possible object-word pairs ${(w_{op}^i, w_{sp}^j)}$ for each object, where $w_{op}^i$ is the word embedding for the category word of the $i$-th object, $w_{sp}^j$ is the word embedding for the $j$-th source token.
Both embeddings are from the same pretrained language model.
Notice that $W_{sp}^n=\{w_{sp}^1,w_{sp}^2,...,w_{sp}^n\}$ is different from $W_s^n$ in that the former is from the pretrained language model and only used for source-relevant object detection in the preprocessing step while the latter is initialized randomly and trained with the model.
We perform max-pooling over the corresponding row of the similarity matrix $S$ to obtain the similarity score of the object to the entire source sentence. In this way, we collect a vector of similarity scores $\textmd{OSS}$ (object-to-sentence similarity) for all objects as follows:
\begin{equation}
\textmd{OSS}_i = \max S_{i,1:n}, ~~i = 1,2,...,m
\end{equation}

We then define an indicator $d$ to indicate whether an object is source-relevant or not as follows:
\begin{equation}
d_i = 1~~ \textmd{if}~~\textmd{OSS}_i>\gamma~~ \textmd{otherwise} ~~0, ~~i = 1,2,...,m
\end{equation}
where $\gamma$ is a predefined similarity threshold hyper-parameter.\footnote{In our case, we set $\gamma$ to 0.48 after randomly checking 100 samples in the training data to select a suitable threshold.}

With $d$, we calculate the object-masking loss as follows:
\begin{equation}
    L_r = L(O^m_{\o_i}, W_{s}^n)~~ \textmd{if}~~ d_i=1
\end{equation}
\begin{equation}
    L_{ir} = L(O^m_{\o_i}, W_{s}^n)~~ \textmd{if}~~ d_i=0
\end{equation}
\begin{equation}
    L_m = -(L_r-L_o) + (L_{ir}-L_o)^2
\end{equation}
where $L$ denotes the cross-entropy translation loss of OVC fed with different visual features, $O^m_{\o_i}$ denotes $O^m$ where the $i$-th object is masked (i.e, $o_i=\textbf{0}$),
$L_o$ denotes the original cross-entropy loss of OVC where no objects are masked,
$L_r$ calculates the new cross-entropy loss if a source-relevant object is masked while $L_{ir}$ is the new loss if a source-irrelevant object is masked.
Therefore, minimizing $L_m$ will force the model to reward masking irrelevant objects and penalize masking relevant objects.
For each training instance, OVC randomly samples source-irrelevant objects for computing $L_{ir}$ and source-relevant objects for generating $L_r$.
For each masked instance, we make sure that all masked objects are either source-relevant or source-irrelevant. No mixed cases are sampled.

\subsubsection{Vision-Weighted Translation Loss.}
Partially inspired by VIFIDEL \cite{vifidel} which checks whether the generated translations are consistent with the visual modality by evaluating the visual fidelity of them, we introduce a vision-weighted translation loss.
Similar to $\textmd{OSS}$, we first compute a target-to-source semantic similarity matrix $S'^{r*n}$ where $r$ is the number of target tokens.
In order to allow the model to pay more attention to vision-related tokens\footnote{Vision-related tokens are marked and provided by Flickr30K-Entities \cite{entities}.} in source texts (e.g., ``man", ``green" in Figure \ref{fig2}), we further set elements that are not vision-related in $S'$ to \textbf{0}.
Then we compute a target-to-vision-related-source similarity vector $\textmd{TVS}$ as follows:
\begin{equation}
    \textmd{TVS}_j = \max S'_{j,1:n}, ~~j = 1,2,...,r
\end{equation}
After that, we calculate a weight for each target word to estimate how much the target word is consistent with the visual modality as follows:
\begin{equation}
    q_j = \frac{\textmd{TVS}_j/f_j} {\sum_{a=1}^r \textmd{TVS}_a/f_a}, ~~ j=1,2,...,r
\end{equation}
where $f_j$ is the frequency of the $j$-th token in the training data. $f_j$ is applied to de-bias rare vision-related words.
Then the vision-weighted loss $L_v$ can be computed as follows:
\begin{equation}
    L_v = \sum_{j=1}^{r} q_{j}*Lo_j
\end{equation}
where $Lo_j$ is the cross-entropy loss of the $j$-th target word.
Generally, $L_v$ favors target words that are vision-consistent.
Rare words can be encouraged to generate if they are related to the visual modality through the de-biasing factor $f_j$.

\subsubsection{Overall Objective of OVC.}
We aggregate the basic translation loss $L_o$, the object-masking loss $L_m$ and the vision-weighted loss $L_v$ for each sample as follows:
\begin{equation}
    L_{ovc} = (Lo+L_{r}+L_{ir})/3 + \alpha*L_m + \beta*L_v
\end{equation}
where $\alpha$ and $\beta$ are two hyper-parameters to control the two additional training objectives.

\section{Experiments}
\label{experiment}
In order to evaluate the proposed OVC framework for MMT, we conducted a series of experiments on MMT datasets and compared with state-of-the-art MMT models.

\subsection{Dataset}
We used three datasets:
\begin{itemize}
    \item Multi30K \cite{multi30k}: This is a widely-used benchmark dataset for MMT, which contains English captions for images from Flickr30K \cite{flicker30k} and corresponding translations into German, French and Czech.
    We conducted experiments with English-to-French (En-Fr) and English-to-German (En-De) and adopted the default split of Multi30K in WMT 2017 MMT shared task, which consists of 29,000 samples for training and 1,014 for validation, and 1,000 for test.
    We used sentences with subwords preprocessed by the implementation of VAG-NMT. For these splits\footnote{\url{https://drive.google.com/drive/folders/1G645SexvhMsLPJhPAPBjc4FnNF7v3N6w}},
    The vocabulary contains 8.5K sub-words for English, 9.4K for German and 8.7K for French.
    \item WMT17 MMT test set \cite{wmt17}: This test set contains 1,000 unduplicated images manually selected from 7 different Flickr groups.
    \item Ambiguous COCO: This is an out-of-domain test set of WMT 2017 with 461 images whose captions are selected to contain ambiguous verbs.
\end{itemize}

\subsection{Experiment Settings}
Following previous works \cite{distilling, gmmt}, we evaluated OVC in the following two settings.
\begin{itemize}
    \item \textbf{Standard setting}: For this setting, we retain all words in source texts and feed them as textual input into all MMT models for both training and inference.
    \item \textbf{Source-degradation setting}: In this setting, we mask words in source texts according to Flickr30K-Entities \cite{entities}, which manually categorizes words in English captions in Multi30K into 9 classes:`people', `scene', `clothing', `instruments', `animals', `bodyparts', `vehicles', `other' and `\texttt{notvisual}'. We did not mask the `\texttt{notvisual}' category as words in this category cannot been grounded in the corresponding image.
        Except for the `\texttt{notvisual}' words, we replaced vision-related words with their corresponding category tags. Besides, we replaced color-related words as an identical `color' category in the remaining source texts, as shown in Figure \ref{fig1}.
        20.9\% of words (79,622 out of 380,793) in the training set and 21.0\% of words (2,818 out of 13,419) in the validation set are masked in this way.
        As Flickr30K-Entities do not provide tags for the re-sampled images in the WMT17 MMT test set, we only evaluated MMT models on the development set in this experiment setting.
        We fed all MMT models with masked source texts as textual input during both training and inference.
\end{itemize}
\begin{table*}[t]
\centering
\begin{tabular}{l|cccc|cccc}
\hline
\multirow{3}{*}{Models}     & \multicolumn{4}{c|}{WMT17 MMT test set}                                                    & \multicolumn{4}{c}{Ambiguous COCO} \\
\cline{2-9}                 & \multicolumn{2}{c}{En$\Rightarrow$Fr} & \multicolumn{2}{c|}{En$\Rightarrow$De}    & \multicolumn{2}{c}{En$\Rightarrow$Fr} & \multicolumn{2}{c}{En$\Rightarrow$De}\\
\cline{2-9}
                                & BLEU          & METEOR        & BLEU          & METEOR        & BLEU          & METEOR        & BLEU  & METEOR\\
\hline
\multicolumn{9}{c}{Existing MMT Models}\\
\hline
($T$) Transformer$\ddag$        & 52.0                  & 68.0          & 30.6                  & 50.4                  & -                     & -                     & 27.3                  & 46.2\\
\cline{2-9}
($R$) Imagination\_i            & -                     &-              & 30.2                  & 51.2                  & -                     & -                     & 26.4                  & 45.8\\
\cline{2-9}
($R$) VAG-NMT\_i$\ddag$         & 53.5$\pm$0.7          & 70.0$\pm$0.7  & 31.6$\pm$0.5          & 52.2$\pm$0.3          & 44.6$\pm$0.6          & 64.2$\pm$0.5          & 27.9$\pm$0.6          & 47.8$\pm$0.6 \\
\cline{2-9}
($R$) VAG-NMT\_i                & 53.8$\pm$0.3          & 70.3$\pm$0.5  & 31.6$\pm$0.3          & 52.2$\pm$0.3          & 45.0$\pm$0.4          & \textbf{64.7$\pm$0.4} & 28.3$\pm$0.6         & 48.0$\pm$0.5 \\
\cline{2-9}
($R$) VAR-MMT\_o                & 52.6                  & 69.9          & 29.3                  & 51.2                  & -                     & -                     & -                     &\\
\cline{2-9}
($T$) VAR-MMT\_o                & 53.3                  & 70.4          & 29.5                  & 50.3                  & -                     & -                     & -                     &\\
\cline{2-9}
($R$) LIUMCVC\_i                & 52.7$\pm$0.9          & 69.5$\pm$0.7  & 30.7$\pm$1.0          & 52.2$\pm$0.4 & 43.5$\pm$1.2          & 63.2$\pm$0.9                      & 26.4$\pm$0.9          & 47.4$\pm$0.3\\
\cline{2-9}
($R$) VMMT\_i                   & -                     & -             & 30.1$\pm$0.3          & 49.9$\pm$0.3          & -                     & -                                 & 25.5$\pm$0.5          & 44.8$\pm$0.2\\
\cline{2-9}
($T$) GMMT\_o                   & 53.9                  & 69.3          & 32.2                  & 51.9                  & -                     & -                                 & 28.7          & 47.6\\
\hline
\multicolumn{9}{c}{Our Proposed Models}\\
\hline
OVC                             & 53.5$\pm$0.2          & 70.2$\pm$0.3          & 31.7$\pm$0.3              & 51.9$\pm$0.4          & 44.7$\pm$0.6              & 64.1$\pm$0.3          & 28.5$\pm$0.5                  & 47.8$\pm$0.3 \\
\cline{2-9}
OVC+$L_m$                       & 54.1$\pm$0.7          & \textbf{70.5$\pm$0.5} & 32.3$\pm$0.6              & \textbf{52.4$\pm$0.3} & \textbf{45.3$\pm$0.5}     & 64.6$\pm$0.5          & \textbf{28.9$\pm$0.5}         & \textbf{48.1$\pm$0.5}\\
\cline{2-9}
OVC+$L_v$                       & \textbf{54.2$\pm$0.4} & 70.5$\pm$0.5          & \textbf{32.4$\pm$0.4}     & 52.3$\pm$0.5          & 45.2$\pm$0.4              & 64.6$\pm$0.3          & 28.6$\pm$0.5                  & 48.0$\pm$0.6\\
\cline{2-9}
OVC+$L_m$+$L_v$                 & 54.0$\pm$0.4          & 70.4$\pm$0.4          & 32.4$\pm$0.6              & 52.2$\pm$0.3          & 45.1$\pm$0.6              & 64.5$\pm$0.5          & 28.8$\pm$0.4                  & 48.0$\pm$0.4\\
\hline
\end{tabular}
\caption{Results of standard experiments. $\ddag$ denotes text-only models. \_i denotes models using image-level features. \_o denotes models using object-level features. $R$ denotes RNN-based approaches. $T$ denotes Transformer-based approaches. $L_m$ is the proposed object masking loss. $L_v$ is the proposed vision-weighted loss.}
\label{standard}
\end{table*}

\subsection{Baselines}
We compared our proposed OVC against 6 different strong baselines:
\begin{itemize}
\item Transformer \cite{transformer}: state-of-the-art neural machine translation architecture with self-attention.
\item Imagination \cite{imagination}: an RNN-based sequence-to-sequence MMT system which implicitly aligns images and their corresponding source texts.
\item VAG-NMT \cite{vag-nmt}: an RNN-/Attention-mixed MMT system using vision-text attention to obtain a vision-aware context representation as the initial state of its decoder.
\item VMMT \cite{latent}: a GRU-based MMT approach that imposes a constraint on the KL term to explore non-negligible mutual information between inputs and a latent variable.
\item GMMT \cite{gmmt}: a stacked graph-based and transformer-based MMT model using object-level features and a textual graph parser for modeling semantic interactions.
\item VAR-MMT \cite{agreement}: an attention-based MMT model that employs visual agreement regularization on visual entity attention via additional word aligners.
\end{itemize}

For fairness, all the models were trained using Multi30K. No extra resource was used.
In the standard setting, we compared OVC against these baselines whose performance on the WMT17 MMT test set are directly reported from their corresponding papers.
Note that the performance of Transformer is taken from \cite{gmmt}.
For the source-degradation setting, we only compared OVC of different objectives as this is a new setting where no results of existing models are available.

\subsection{Results in the Standard Setting}
\begin{table}[t]
\centering
\begin{tabular}{|l|c|c|}
\hline
 \multicolumn{3}{|c|}{En$\Rightarrow$De}  \\
\hline
Metrics                 & BLEU  & METEOR   \\
\hline
OVC\_t                  & 21.02 & 40.61 \\
\hline
OVC\_i                  & 22.02 & 41.91  \\
\hline
OVC\_o                  & 21.98 & 41.57 \\
\hline
OVC\_o+$HM$             & 25.31 & 43.85 \\
\hline
OVC\_o+$L_m$            & 26.30 & 45.37 \\
\hline
OVC\_o+$L_v$            & 22.18 & 42.01 \\
\hline
OVC\_o+$L_m$+$L_v$      & 22.57 & 42.24 \\
\hline
\multicolumn{3}{|c|}{En$\Rightarrow$Fr}\\
\hline
OVC\_t                  & 37.01 & 55.35 \\
\hline
OVC\_i                  & 37.40 & 55.68 \\
\hline
OVC\_o                  & 36.94 & 54.92 \\
\hline
OVC\_o+$HM$             & 37.39 & 55.38 \\
\hline
OVC\_o+$L_m$            & 39.31 & 57.28 \\
\hline
OVC\_o+$L_v$            & 37.25 & 55.79 \\
\hline
OVC\_o+$L_m$+$L_v$      & 37.63 & 56.14 \\
\hline
\end{tabular}
\caption{Results for the source-degradation setting on the WMT17 MMT development set. \_t denotes text-only models. $HM$ denotes a hard masking scheme where irrelevant objects are masked in a hard way via the pretrained threshold.}
\label{source-degradation}
\end{table}

\subsection{Model Setting for OVC}
In order to avoid the influence of the increasing number of parameters on the comparison, we limited the number of parameters in our OVC models to be comparative to that in \cite{vag-nmt} (16.0M parameters).
In order to achieve this, we set the size of word embeddings in OVC to 256.
The encoder of source texts has one bidirectional-GRU layer and one multihead object-text attention layer.
The hidden state sizes of all modules in the encoder were set to 512.
The decoder has one multihead attention layer and two stacked GRU layers, of which the hidden sizes were set to 512 and the input sizes 256 and 512 for the two GRU layers, respectively.
We used Adam as the optimizer with a scheduled learning rate and applied early-stopping with a patient step of 10 during training.
With these settings, our proposed OVC of its full form has 11.3M parameters.
All models were trained in the teacher-forcing manner.
Other settings were kept the same as in \cite{vag-nmt}.
All implementations were built based upon Pytorch and models were both trained and evaluated on one 2080Ti GPU.
We performed a grid search on the WMT17 MMT development set to obtain the hyper-parameters: $\alpha$ was set to 0.1 and $\beta$ was set to 0.1.

For image-level visual features, we used the pool5 outputs of a pretrained Resnet-50, released by WMT 2017.
For object-level visual features, we first took the pool5 outputs of a pretrained Resnet101 detector\footnote{\url{https://github.com/peteanderson80/bottom-up-attention}} as candidates. We then selected objects of the highest 20 object confidences as our object-level features.

To make our experiments more statistically reliable, for the proposed model, we run each experiment for three times and report the average results over the three runs.
The results in the standard setting are listed in Table \ref{standard}.
OVC trained with the two additional losses either outperforms existing Transformer-based and RNN-based MMT models with an average improvement of 0.25 BLEU and 0.10 METEOR, or achieves competitive results to them.
The basic OVC shows no advantage over existing image-level MMT models. For example, in most cases, the basic OVC is not better than VAG-NMT\_i on the WMT17 MMT test set and Ambiguous COCO.
We conjecture that the object-level visual features may contain irrelevant information for machine translation. And since the Multi30K training data is small and textually repetitive, this makes it hard for object-level MMT models to learn fine-grained grounding alignments.
However, after being equipped with the two proposed additional objectives, OVC is superior to both image- and object-level MMT models.
It gains an average improvement of 0.4$\sim$0.6 BLEU and 0.3$\sim$0.5 METEOR using the additional $L_m$, while 0.1$\sim$0.7 BLEU and 0.2$\sim$0.5 METEOR using the additional $L_v$, which indicate that our proposed objectives enhance the visual grounding capability of OVC.
Additionally, we visualize the object-source attention of OVC trained with different objectives in the Appendix to support this hypothesis.

\subsection{Results in Source-Degradation Setting and Ablation Study}

In this setting, we compared different OVC variants using different objectives, which is also the ablation study of our proposed OVC.
We also trained OVC in a text-only setting by dropping the object-to-source attention layer in its encoder, where VASR is replaced by the annotation vectors and SSV is directly the average-pooling result of the annotation vectors.

The results are shown in Table \ref{source-degradation}. Under the source-degradation setting, with image-level features, OVC is better than its text-only version, which is consistent with previous multimodal machine translation findings \cite{probing}.
With object-level features, the performance of OVC is generally worse than that with image-level features and even worse than the text-only OVC on English-to-French translation.
This again confirms our finding with the basic OVC under the standard setting.
Besides, it can be seen that the improvements of both $L_m$ and $L_v$ in the source-degradation setting are generally larger than those in the standard setting.
Particularly, $L_m$ gains an average improvement of 3.35 BLEU and 3.08 METEOR while $L_v$ achieves an average improvement of 0.255 BLEU of 0.655 METEOR over the basic OVC.

For a deep understanding on the impact of object masking, we further compared a hard masking scheme where source-irrelevant objects are compulsively masked in a hard way instead of using the training objective in a soft way according to the predefined similarity threshold. The stable improvement of behavior of OVC\_o+$HM$ vs. OVC\_o and OVC\_o+$L_{m}$ vs. OVC\_o+$HM$ suggest that masking irrelevant objects helps grounding in MMT as vision-related words are all masked in the degraded source sentences.
Since the only difference between $L_m$ and $HM$ is that $L_m$ penalizes masking source-relevant objects and encourages masking source-irrelevant objects simultaneously in a soft way, the improvements of $L_m$ over $HM$ indicate that the proposed object-masking loss is a more efficient way for grounding in MMT.
\begin{table}[t]
\centering
\begin{tabular}{|c|c|c|}
\hline
ST:SD           & BLEU  & METEOR \\
\hline
1.0:0.0         & 22.43 & 41.64 \\
\hline
1.0:0.2         & 22.66 & 41.53 \\
\hline
1.0:0.4         & \textbf{23.01} & \textbf{42.08} \\
\hline
1.0:0.5         & 22.75 & 41.73 \\
\hline
1.0:0.6         & 22.68 & 41.68 \\
\hline
1.0:0.8         & 22.82 & 42.00 \\
\hline
1.0:1.0         & 22.05 & 41.03 \\
\hline
\end{tabular}
\caption{Results under the mixed setting on the WMT17 MMT En$\Rightarrow$De development set. ST denotes the number of standard samples while SD denotes the number of source-degradation samples.}
\label{mix}
\end{table}
\section{Analysis}

\subsection{Case Analysis}

Apart from the visualization of the attention of OVC in different model settings, we also randomly selected samples in the evaluation data to analyze the behavior of different OVC variants on source-degradation samples.
\subsection{Mixed Setting}
Finally, we trained MMT models in a mixed setting where source-degradation and standard texts are mixed together for training and evaluation is done on the source-degradation data.
Specifically, we trained OVC with the source-degradation \& standard mixed training set of Multi30K and evaluated it on the source-degradation samples of the WMT17 MMT En$\Rightarrow$De development set to investigate the potential ability of the source-degraded framework in helping standard MMT.
The results are shown in Table \ref{mix} with different proportions of mixed standard samples and degraded samples.

It is interesting to find that the performance of OVC does not consistently rise as the number of sampled source-degradation samples increase.
The best proportion of additional source-degradation data is 1.0:0.4.
We assume that a certain amount of source-degradation samples can improve the grounding ability of MMT models, which offsets the information loss in source-degradation samples.
However, more source-degradation sample may undermine the ability of MMT in conveying the meaning of source sentences to target translations.

\section{Conclusion and Future Work}

In this paper, to efficiently model the language-vision interaction and integrate visual context into multimodal machine translation, we have presented OVC, an object-level visual context modeling framework.
In OVC, we model the interaction between the textual and visual modality through the object-text similarity and object-source multihead attention on the source side as well as the vision-weighted loss on the target side.
In order to tailor the visual feature learning for multimodal machine translation, the additional object-masking loss is proposed to force OVC to be aware of whether the masked objects are relevant to source texts and to perform desirable masking in a soft way.
The presented vision-weighted translation loss is to guide the decoder to generate vision-consistent target words.
Experiment results show that our proposed framework achieves competitive performance against several existing state-of-the-art MMT models in the standard setting.
 Experiments and analyses on the source-degradation settings suggest that the proposed two additional training objectives, especially the object-masking loss, helps grounding in MMT.

 In the future, we plan to improve the proposed OVC in grounding via other mechanisms (e.g., cross-modality pretraining).
And we are also interested in extending our OVC framework to the video-guided MMT \cite{videamt}.

\begin{figure*}[t]
 \centering
 \includegraphics[width=2.1\columnwidth]{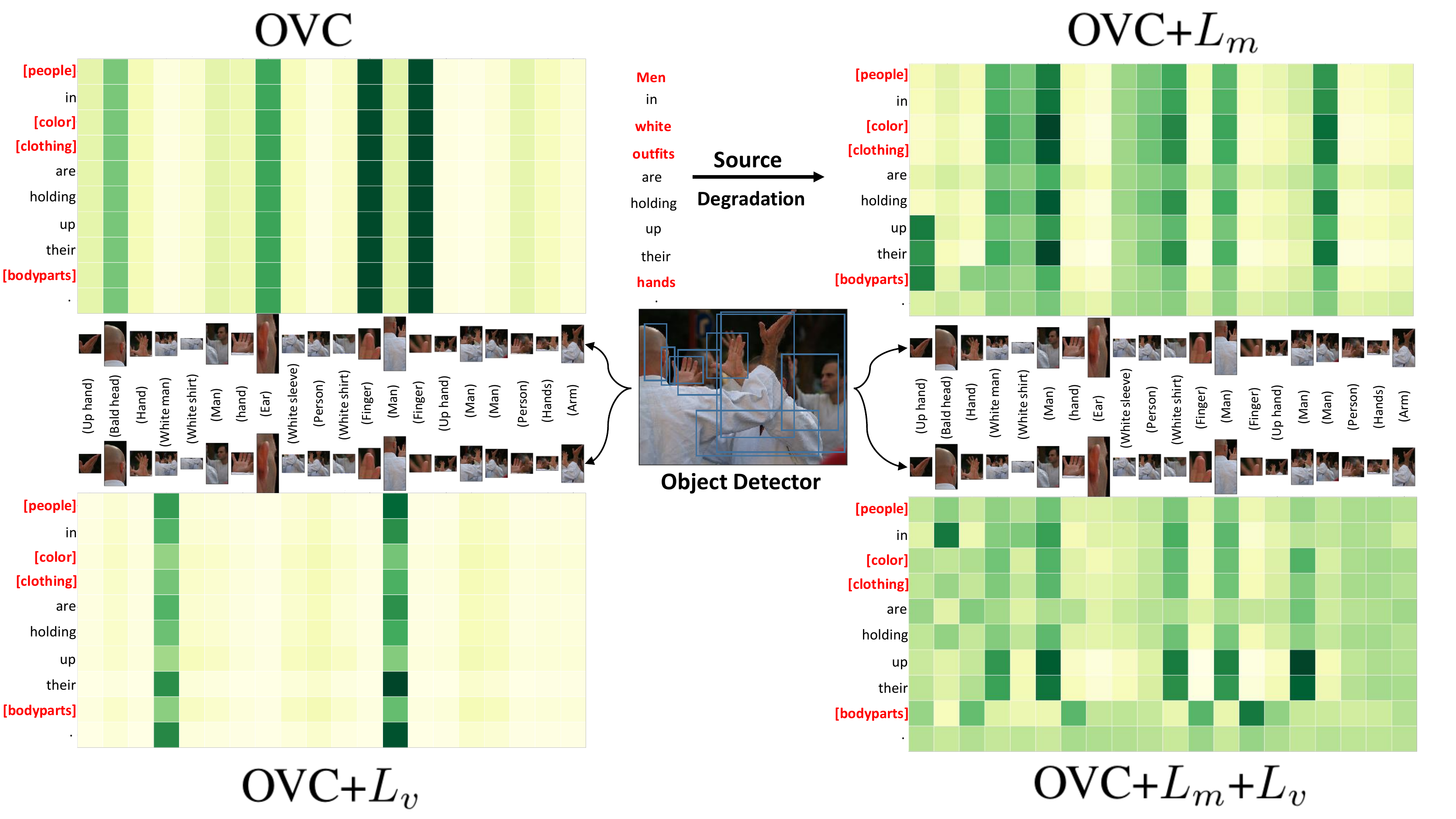} 
 \caption{A source-degraded example from the WMT17 MMT EN$\Rightarrow$DE development set to visualize the source-object attention of OVC variants using the degraded English text as the source text. For better visualization and understanding the attention results, we show the corresponding object category, predicted by the Object Detector, of each detected object in the middle of two parallel rows of objects.}
 \label{object-source-attention}
\end{figure*}

\section{Acknowledgments}
The present research was supported by the National Key Research and Development Program of China (Grant No. 2019QY1802). We would like to thank the anonymous reviewers for their insightful comments. The corresponding author is Deyi Xiong (dyxiong@tju.edu.cn).

\section{Appendix}

\begin{table*}[t]
\centering
\begin{tabular}{c|l}
\hline
Images    & \multicolumn{1}{c}{Descriptions} \\
\hline
\multirow{7}{*}{
\begin{minipage}[b]{0.35\columnwidth}
	\centering
	{\includegraphics[width=\linewidth]{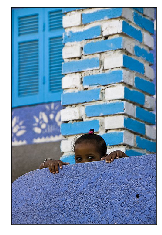}}
\end{minipage}
}
& SRC:  ~~~a little girl peering over a blue wall . \\
& DSRC:  a little [people] peering over a [color] wall . \\
& REF:  ~~~ein kleines m$\ddot{a}$dchen sp$\ddot{a}$ht $\ddot{u}$ber eine blaue mauer .\\
& OVC:    ~     ein kleiner \underline{junge} blickt $\ddot{u}$ber eine \underline{gr$\ddot{u}$ne} wand . \\
& ~~~~~~~~~~~~ (a little boy looks over a green wall .)\\
& OVC+$L_m$:    ein kleiner \underline{junge} guckt $\ddot{u}$ber eine \underline{wei${\ss}$e} wand .   \\
& ~~~~~~~~~~~~ (a little boy looks over a white wall .)\\
& OVC+$L_v$:    ein kleiner \textbf{m$\ddot{a}$dchen} guckt $\ddot{u}$ber eine \underline{wei${\ss}$e} wand .  \\
& ~~~~~~~~~~~~ (a little girl looks over a white wall .)\\
& OVC+$L_m$+$L_v$:  ein kleines \textbf{m$\ddot{a}$dchen} guckt $\ddot{u}$ber eine \textbf{blaue} wand .  \\
& ~~~~~~~~~~~~ (a little girl looks over a blue wall .)\\
\hline

\multirow{7}{*}{
\begin{minipage}[b]{0.66\columnwidth}
	\centering
    \raisebox{-1.\height}
	{\includegraphics[width=\linewidth]{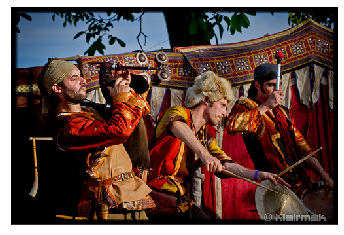}}
\end{minipage}
}
& SRC: ~~~a group of men in costume play music .  \\
& DSRC: a group of [people] in [clothing] play music .  \\
& REF: ~~~eine gruppe von m$\ddot{a}$nnern in kost$\ddot{u}$men spielt musik . \\
& OVC:   ~       eine gruppe von \underline{kindern} in \textbf{kost$\ddot{u}$men} spielt musik .  \\
& ~~~~~~~~~~~~ (a group of children in costumes play music .)\\
& OVC+$L_m$:    eine gruppe von \textbf{m$\ddot{a}$nnern} in \underline{uniform} spielt musik .   \\
& ~~~~~~~~~~~~ (a group of men in uniform plays music .)\\
& OVC+$L_v$:    eine gruppe von \textbf{m$\ddot{a}$nnern} in \underline{anz$\ddot{u}$gen} macht musik .   \\
& ~~~~~~~~~~~~ (a group of men in suits makes music .)\\
& OVC+$L_m$+$L_v$:  eine gruppe von \textbf{m$\ddot{a}$nnern} in \textbf{kost$\ddot{u}$men} spielt musik .   \\
& ~~~~~~~~~~~~ (a group of men in costumes is playing music .)\\
\hline

\multirow{7}{*}{
\begin{minipage}[b]{0.66\columnwidth}
	\centering
    \raisebox{-1.\height}
	{\includegraphics[width=\linewidth]{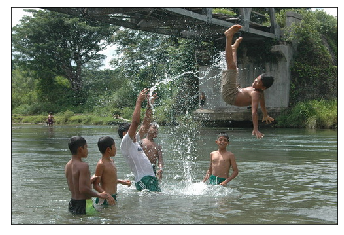}}
\end{minipage}
}
& SRC: ~~~a group of children play in the water under a bridge .   \\
& DSRC: a group of [people] play in the [scene] under a [scene] .   \\
& REF: ~~~eine gruppe von kindern spielt im wasser unter einer br$\ddot{u}$cke . \\
& OVC:   ~      eine gruppe von \textbf{kindern} spielt im \underline{gras} unter einem \underline{berg} . \\
& ~~~~~~~~~~~~ (a group of children play in the grass under a mountain .)\\
& OVC+$L_m$:    eine gruppe \textbf{kinder} spielt im \textbf{wasser} unter einem \underline{wasserfall} .   \\
& ~~~~~~~~~~~~ (a group of children play in the water under a waterfall .)\\
& OVC+$L_v$:    eine gruppe \textbf{kinder} spielt im \textbf{wasser} unter einem \underline{wasserfall} .  \\
& ~~~~~~~~~~~~ (a group of children play in the water under a waterfall .)\\
& OVC+$L_m$+$L_v$:  eine gruppe von \textbf{kindern} spielt im \underline{schnee} unter einem \textbf{br$\ddot{u}$cke} .  \\
& ~~~~~~~~~~~~ (a group of children play in the snow under a bridge .)\\
\hline
\end{tabular}
\caption{Example translations of OVC variants from the source-degradation WMT17 MMT EN$\Rightarrow$DE development set. Each target translation is accompanied with an English re-translation for easy understanding. SRC: the source text. DSRC: the source-degradation source text. REF: the reference translation. Correct translations of the degraded tokens are bold while inappropriately translated tokens are underlined.}
\label{case analyse}
\end{table*}

\subsection{Grounding Capability of OVC in Terms of Source-Object Attention}
To investigate that the grounding capability of our proposed OVC for MMT is enhanced by the new training objectives (object-masking objective $L_m$ and vision-weighted translation ojective $L_v$),
we randomly sampled an sample from source-degradation data derived from the WMT17 MMT development set and visualized the object-source attention of OVC to illustrate the grounding differences between OVC variants equipped with different objectives, as shown in Figure \ref{object-source-attention}.
Each grid represents the attention score of a detected object and a source token.
The sum of each row of attention scores is equal to 1.
The darker a cell of the visualized attention matrix is, the larger the attention score of its corresponding visual object and source token is.
It can be seen that the base OVC tends to assign unreasonably large attention scores to 4 translation-irrelevant objects (``Bald head", ``Ear" and two ``Fingers" in the given image).
Additionally, all cells in each column of the 4 objects are painted with similar colors, which suggests that each of these 4 objects has almost the same scores of attention to different words in the source-degradation text.
We conjecture that the base OVC may be over-fitting to visual object features and hence not capable of appropriately grounding source-object relations for MMT in this case.
OVC+$L_v$ partly improves the grounding to source-relevant objects (``White man" and ``Man"), while some degraded words (e.g., [color] and [bodyparts]) are not yet grounded on visual objects.
The reason may lie in that the training set is small and the degradation of the [people] category occurs much more frequently than those of the [clothing] and [bodyparts] categories in the source-degraded training set.
Hence, a larger vision-weighted loss is imposed on translating people-relevant words, which forces OVC to naturally assign much smaller attention scores to irrelevant objects.

The object-masking loss helps OVC to learn which objects are irrelevant to the source text, hence improving the grounding of more details in the text on the corresponding objects (e.g., the [bodyparts] token has a much larger attention score with the ``Up hand" object). 
Finally, OVC equipped with the joint $L_m$ and $L_v$ obtains the best grounding result in terms of the source-object attention among the four variants of OVC.

\subsection{Case Analysis of OVC on Source-Degradation Samples}

For case analysis, we randomly sampled data from WMT17 MMT development set and converted them into their source-degradation version to compare MMT translations, as shown in Table \ref{case analyse}.
In these cases, the proposed vision-weighted loss and object-masking loss improve the translation for degraded `gender', `color', `clothing' and `scene' categories.
However, we also find it hard to adequately translate a short source sentence with multiple degraded tokens from the same category. This is not only related to the challenging visual grounding problem in MMT, but also to the difficulty in finding the right combination of potential object candidates and aligning them to degraded source tokens in the same category.

\bibliography{bibtex}

\begin{thebibliography}{21}
\providecommand{\natexlab}[1]{#1}
\providecommand{\url}[1]{\texttt{#1}}
\providecommand{\urlprefix}{URL }
\expandafter\ifx\csname urlstyle\endcsname\relax
  \providecommand{\doi}[1]{doi:\discretionary{}{}{}#1}\else
  \providecommand{\doi}{doi:\discretionary{}{}{}\begingroup
  \urlstyle{rm}\Url}\fi

\bibitem[{Caglayan et~al.(2017)Caglayan, Aransa, Bardet,
  Garc{\'\i}a-Mart{\'\i}nez, Bougares, Barrault, Masana, Herranz, and van~de
  Weijer}]{LIUMCVC}
Caglayan, O.; Aransa, W.; Bardet, A.; Garc{\'\i}a-Mart{\'\i}nez, M.; Bougares,
  F.; Barrault, L.; Masana, M.; Herranz, L.; and van~de Weijer, J. 2017.
\newblock {LIUM}-{CVC} Submissions for {WMT}17 Multimodal Translation Task.
\newblock In \emph{Proceedings of the Second Conference on Machine
  Translation}, 432--439. Copenhagen, Denmark: Association for Computational
  Linguistics.
\newblock \doi{10.18653/v1/W17-4746}.
\newblock \urlprefix\url{https://www.aclweb.org/anthology/W17-4746}.

\bibitem[{Caglayan et~al.(2019)Caglayan, Madhyastha, Specia, and
  Barrault}]{probing}
Caglayan, O.; Madhyastha, P.; Specia, L.; and Barrault, L. 2019.
\newblock Probing the Need for Visual Context in Multimodal Machine
  Translation.
\newblock In \emph{Proceedings of the 2019 Conference of the North {A}merican
  Chapter of the Association for Computational Linguistics: Human Language
  Technologies, Volume 1 (Long and Short Papers)}, 4159--4170. Minneapolis,
  Minnesota: Association for Computational Linguistics.
\newblock \doi{10.18653/v1/N19-1422}.
\newblock \urlprefix\url{https://www.aclweb.org/anthology/N19-1422}.

\bibitem[{Calixto, Rios, and Aziz(2019)}]{latent}
Calixto, I.; Rios, M.; and Aziz, W. 2019.
\newblock Latent Variable Model for Multi-modal Translation.
\newblock In \emph{Proceedings of the 57th Annual Meeting of the Association
  for Computational Linguistics}, 6392--6405. Florence, Italy: Association for
  Computational Linguistics.
\newblock \doi{10.18653/v1/P19-1642}.
\newblock \urlprefix\url{https://www.aclweb.org/anthology/P19-1642}.

\bibitem[{Elliott et~al.(2017)Elliott, Frank, Barrault, Bougares, and
  Specia}]{wmt17}
Elliott, D.; Frank, S.; Barrault, L.; Bougares, F.; and Specia, L. 2017.
\newblock Findings of the Second Shared Task on Multimodal Machine Translation
  and Multilingual Image Description.
\newblock In \emph{Proceedings of the Second Conference on Machine
  Translation}, 215--233. Copenhagen, Denmark: Association for Computational
  Linguistics.
\newblock \doi{10.18653/v1/W17-4718}.
\newblock \urlprefix\url{https://www.aclweb.org/anthology/W17-4718}.

\bibitem[{Elliott et~al.(2016)Elliott, Frank, Sima{'}an, and Specia}]{multi30k}
Elliott, D.; Frank, S.; Sima{'}an, K.; and Specia, L. 2016.
\newblock {M}ulti30{K}: Multilingual {E}nglish-{G}erman Image Descriptions.
\newblock In \emph{Proceedings of the 5th Workshop on Vision and Language},
  70--74. Berlin, Germany: Association for Computational Linguistics.
\newblock \doi{10.18653/v1/W16-3210}.
\newblock \urlprefix\url{https://www.aclweb.org/anthology/W16-3210}.

\bibitem[{Elliott and K{\'a}d{\'a}r(2017)}]{imagination}
Elliott, D.; and K{\'a}d{\'a}r, {\'A}. 2017.
\newblock Imagination Improves Multimodal Translation.
\newblock In \emph{Proceedings of the Eighth International Joint Conference on
  Natural Language Processing (Volume 1: Long Papers)}, 130--141. Taipei,
  Taiwan: Asian Federation of Natural Language Processing.
\newblock \urlprefix\url{https://www.aclweb.org/anthology/I17-1014}.

\bibitem[{He et~al.(2016)He, Zhang, Ren, and Sun}]{resnet}
He, K.; Zhang, X.; Ren, S.; and Sun, J. 2016.
\newblock Deep residual learning for image recognition.
\newblock In \emph{Proceedings of the IEEE conference on computer vision and
  pattern recognition}, 770--778.

\bibitem[{Hirasawa et~al.(2019)Hirasawa, Yamagishi, Matsumura, and
  Komachi}]{embedding}
Hirasawa, T.; Yamagishi, H.; Matsumura, Y.; and Komachi, M. 2019.
\newblock Multimodal Machine Translation with Embedding Prediction.
\newblock In \emph{Proceedings of the 2019 Conference of the North {A}merican
  Chapter of the Association for Computational Linguistics: Student Research
  Workshop}, 86--91. Minneapolis, Minnesota: Association for Computational
  Linguistics.
\newblock \doi{10.18653/v1/N19-3012}.
\newblock \urlprefix\url{https://www.aclweb.org/anthology/N19-3012}.

\bibitem[{Huang et~al.(2016)Huang, Liu, Shiang, Oh, and Dyer}]{huang}
Huang, P.-Y.; Liu, F.; Shiang, S.-R.; Oh, J.; and Dyer, C. 2016.
\newblock Attention-based Multimodal Neural Machine Translation.
\newblock In \emph{Proceedings of the First Conference on Machine Translation:
  Volume 2, Shared Task Papers}, 639--645. Berlin, Germany: Association for
  Computational Linguistics.
\newblock \doi{10.18653/v1/W16-2360}.
\newblock \urlprefix\url{https://www.aclweb.org/anthology/W16-2360}.

\bibitem[{Ive, Madhyastha, and Specia(2019)}]{distilling}
Ive, J.; Madhyastha, P.; and Specia, L. 2019.
\newblock Distilling Translations with Visual Awareness.
\newblock In \emph{Proceedings of the 57th Annual Meeting of the Association
  for Computational Linguistics}, 6525--6538. Florence, Italy: Association for
  Computational Linguistics.
\newblock \doi{10.18653/v1/P19-1653}.
\newblock \urlprefix\url{https://www.aclweb.org/anthology/P19-1653}.

\bibitem[{Lala et~al.(2018)Lala, Madhyastha, Scarton, and Specia}]{sheffield}
Lala, C.; Madhyastha, P.~S.; Scarton, C.; and Specia, L. 2018.
\newblock {S}heffield Submissions for {WMT}18 Multimodal Translation Shared
  Task.
\newblock In \emph{Proceedings of the Third Conference on Machine Translation:
  Shared Task Papers}, 624--631. Belgium, Brussels: Association for
  Computational Linguistics.
\newblock \doi{10.18653/v1/W18-6442}.
\newblock \urlprefix\url{https://www.aclweb.org/anthology/W18-6442}.

\bibitem[{Lee et~al.(2018)Lee, Cho, Weston, and Kiela}]{emergent}
Lee, J.; Cho, K.; Weston, J.; and Kiela, D. 2018.
\newblock Emergent Translation in Multi-Agent Communication.
\newblock In \emph{Proceedings of the International Conference on Learning
  Representations}.

\bibitem[{Madhyastha, Wang, and Specia(2019)}]{vifidel}
Madhyastha, P.; Wang, J.; and Specia, L. 2019.
\newblock {VIFIDEL}: Evaluating the Visual Fidelity of Image Descriptions.
\newblock In \emph{Proceedings of the 57th Annual Meeting of the Association
  for Computational Linguistics}, 6539--6550. Florence, Italy: Association for
  Computational Linguistics.
\newblock \doi{10.18653/v1/P19-1654}.
\newblock \urlprefix\url{https://www.aclweb.org/anthology/P19-1654}.

\bibitem[{{Plummer} et~al.(2015){Plummer}, {Wang}, {Cervantes}, {Caicedo},
  {Hockenmaier}, and {Lazebnik}}]{entities}
{Plummer}, B.~A.; {Wang}, L.; {Cervantes}, C.~M.; {Caicedo}, J.~C.;
  {Hockenmaier}, J.; and {Lazebnik}, S. 2015.
\newblock Flickr30k Entities: Collecting Region-to-Phrase Correspondences for
  Richer Image-to-Sentence Models.
\newblock In \emph{2015 IEEE International Conference on Computer Vision
  (ICCV)}, 2641--2649.

\bibitem[{Vaswani et~al.(2017)Vaswani, Shazeer, Parmar, Uszkoreit, Jones,
  Gomez, Kaiser, and Polosukhin}]{transformer}
Vaswani, A.; Shazeer, N.; Parmar, N.; Uszkoreit, J.; Jones, L.; Gomez, A.~N.;
  Kaiser, L.; and Polosukhin, I. 2017.
\newblock Attention is All you Need.
\newblock \emph{ArXiv} abs/1706.03762.

\bibitem[{Wang et~al.(2019)Wang, Wu, Chen, Li, Wang, and Wang}]{videamt}
Wang, X.; Wu, J.; Chen, J.; Li, L.; Wang, Y.; and Wang, W.~Y. 2019.
\newblock VaTeX: A Large-Scale, High-Quality Multilingual Dataset for
  Video-and-Language Research.
\newblock \emph{2019 IEEE/CVF International Conference on Computer Vision
  (ICCV)} 4580--4590.

\bibitem[{Yang et~al.(2020)Yang, Chen, Zhang, and Sun}]{agreement}
Yang, P.; Chen, B.; Zhang, P.; and Sun, X. 2020.
\newblock Visual Agreement Regularized Training for Multi-Modal Machine
  Translation.
\newblock In \emph{AAAI}, 9418--9425.

\bibitem[{Yin et~al.(2020)Yin, Meng, Su, Zhou, Yang, Zhou, and Luo}]{gmmt}
Yin, Y.; Meng, F.; Su, J.; Zhou, C.; Yang, Z.; Zhou, J.; and Luo, J. 2020.
\newblock A Novel Graph-based Multi-modal Fusion Encoder for Neural Machine
  Translation.
\newblock In \emph{Proceedings of the 58th Annual Meeting of the Association
  for Computational Linguistics}, 3025--3035. Online: Association for
  Computational Linguistics.
\newblock \doi{10.18653/v1/2020.acl-main.273}.
\newblock \urlprefix\url{https://www.aclweb.org/anthology/2020.acl-main.273}.

\bibitem[{Young et~al.(2014)Young, Lai, Hodosh, and Hockenmaier}]{flicker30k}
Young, P.; Lai, A.; Hodosh, M.; and Hockenmaier, J. 2014.
\newblock From image descriptions to visual denotations: New similarity metrics
  for semantic inference over event descriptions.
\newblock \emph{Transactions of the Association for Computational Linguistics}
  2: 67--78.
\newblock \doi{10.1162/tacl_a_00166}.
\newblock \urlprefix\url{https://www.aclweb.org/anthology/Q14-1006}.

\bibitem[{Zhang et~al.(2020)Zhang, Chen, Wang, Utiyama, Sumita, Li, and
  Zhao}]{zhang}
Zhang, Z.; Chen, K.; Wang, R.; Utiyama, M.; Sumita, E.; Li, Z.; and Zhao, H.
  2020.
\newblock Neural Machine Translation with Universal Visual Representation.
\newblock In \emph{ICLR}.

\bibitem[{Zhou et~al.(2018)Zhou, Cheng, Lee, and Yu}]{vag-nmt}
Zhou, M.; Cheng, R.; Lee, Y.~J.; and Yu, Z. 2018.
\newblock A Visual Attention Grounding Neural Model for Multimodal Machine
  Translation.
\newblock In \emph{Proceedings of the 2018 Conference on Empirical Methods in
  Natural Language Processing}, 3643--3653. Brussels, Belgium: Association for
  Computational Linguistics.
\newblock \doi{10.18653/v1/D18-1400}.
\newblock \urlprefix\url{https://www.aclweb.org/anthology/D18-1400}.

\end{thebibliography}

\end{document}